\documentclass[10pt,twocolumn,letterpaper]{article}

\usepackage{cvpr}              % To produce the CAMERA-READY version
\usepackage{graphicx}
\usepackage{amsmath}
\usepackage{amssymb}
\usepackage{booktabs}
\usepackage{multirow}
\usepackage{makecell}
\usepackage{threeparttable}
\usepackage{lipsum}
\usepackage{tabularx}
\usepackage{diagbox}

\usepackage[pagebackref,breaklinks,colorlinks]{hyperref}

\usepackage[capitalize]{cleveref}
\crefname{section}{Sec.}{Secs.}
\Crefname{section}{Section}{Sections}
\Crefname{table}{Table}{Tables}
\crefname{table}{Tab.}{Tabs.}

\def\cvprPaperID{2676} % *** Enter the CVPR Paper ID here
\def\confName{CVPR}
\def\confYear{2022}

\begin{document}

%%%%%%%%% TITLE - PLEASE UPDATE
\title{Distillation with Contrast is All You Need for Self-Supervised Point Cloud Representation Learning}

\author{Kexue Fu$^{1,2}$ \ \ \ Peng Gao$^{2}$ \ \ \ Renrui Zhang$^{2}$ \ \ \ Hongsheng Li$^{3}$ \ \ \ Yu Qiao$^{2,4}$ \ \ \ Manning Wang$^{1}$\thanks{Corresponding author}\\
1 Digital Medical Research Center, School of Basic Medical Science, Fudan University \\
2 Shanghai AI Lab \\
3 The Chinese University of Hong Kong \\
4 Shenzhen Institutes of Advanced Technology, Chinese Academy of Sciences\\
% For a paper whose authors are all at the same institution,
% omit the following lines up until the closing ``}''.
% Additional authors and addresses can be added with ``\and'',
% just like the second author.
% To save space, use either the email address or home page, not both
{\tt $\{$kxfu18, mnwang$\}$@fudan.edu.cn }
}
\maketitle

%%%%%%%%% ABSTRACT
\begin{abstract}
  In this paper, we propose a simple and general framework for self-supervised point cloud representation learning. Human beings understand the 3D world by extracting two levels of information and establishing the relationship between them. One is the global shape of an object, and the other is the local structures of it. However, few existing studies in point cloud representation learning explored how to learn both global shapes and local-to-global relationships without a specified network architecture. Inspired by how human beings understand the world, we utilize knowledge distillation to learn both global shape information and the relationship between global shape and local structures. At the same time, we combine contrastive learning with knowledge distillation to make the teacher network be better updated. Our method achieves the state-of-the-art performance on linear classification and multiple other downstream tasks. Especially, we develop a variant of ViT for 3D point cloud feature extraction, which also achieves comparable results with existing backbones when combined with our framework, and visualization of the attention maps show that our model does understand the point cloud by combining the global shape information and multiple local structural information, which is consistent with the inspiration of our representation learning method. Our code will be released soon.
\end{abstract}

%%%%%%%%% BODY TEXT
\section{Introduction}
\label{sec:intro}
% figure 1
\begin{figure}[ht]
  \centering
  \setlength{\belowcaptionskip}{-0.5cm}
  \includegraphics[width=7.5cm]{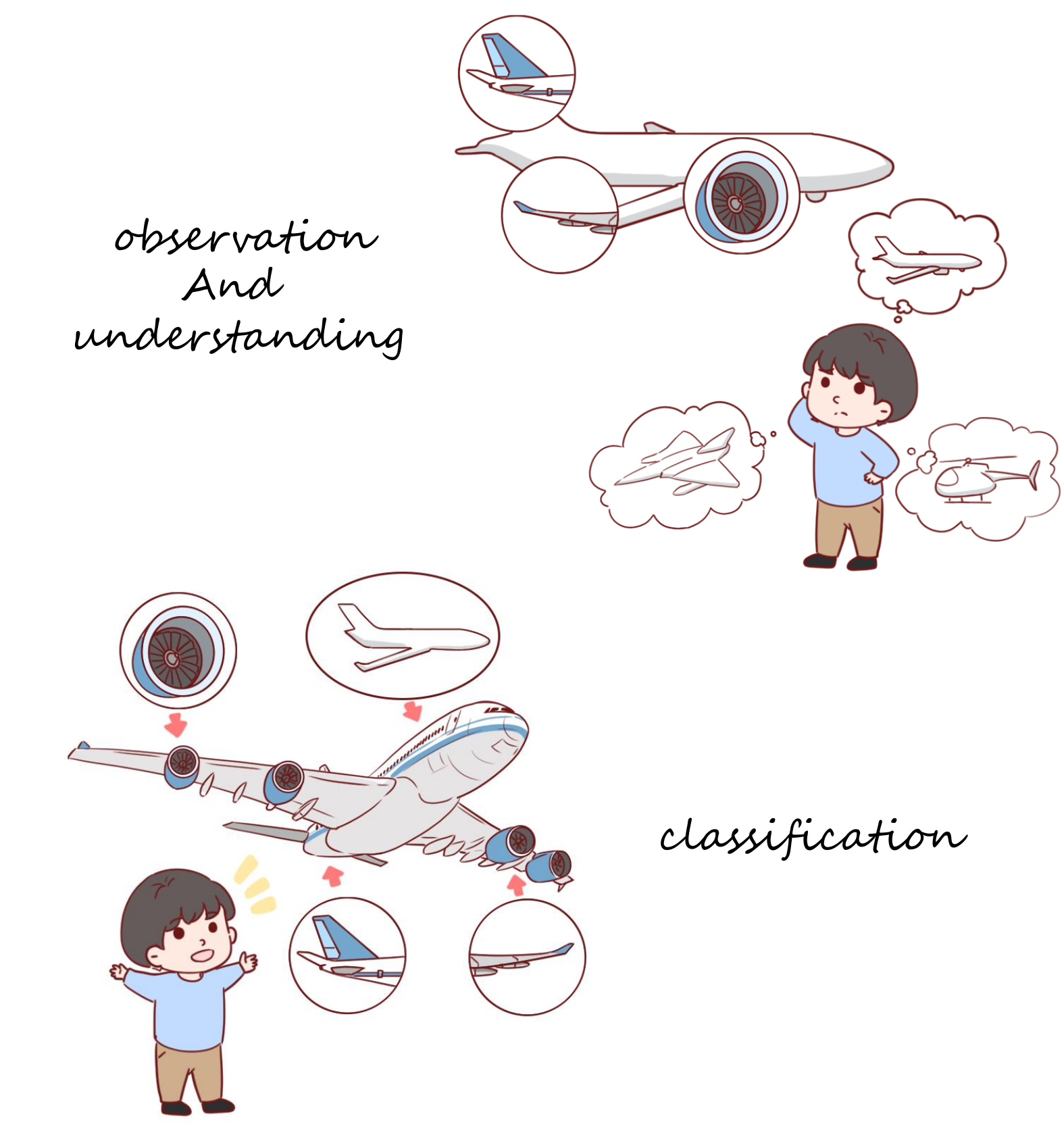} 
  \caption{Illustration of our inspiration and main idea. The top figure shows how a child learns to understand the airplane. He first remembers the vague outline of the airplane, and then observes the key parts of it. When he sees a new object as shown in the bottom figure, he will match the outline with his memory and observe whether there are some key parts to determine whether it is an airplane.}
  \label{fig1}
\end{figure}

Point cloud is an intuitive, flexible and memory-efficient 3D data representation and has become indispensable in 3D vision. Learning powerful point cloud representation is very crucial for facilitating machines to understand the 3D world, which is beneficial for promoting the development of many important real-world applications, such as autonomous driving \cite{ref28}, augmented reality \cite{ref30} and robotics \cite{ref29}. With the rapid development of deep learning in these years \cite{ref43,ref44}, supervised 3D point cloud analysis methods have made great progress \cite{ref19,ref20,ref21}. However, both exponentially increasing demand for data and expensive 3D data annotation hinder further performance improvement of supervised methods. On the contrary, due to the widespread popularity of 3D sensors (Lidar, ToF camera, RGB-D sensor or camera stereo-pair), a large number of unlabeled point cloud data are available for self-supervised point cloud representation learning.

Unsupervised or self-supervised learning methods have shown their effectiveness in different fields \cite{method3, ref23, method2, ref26, method11}. In the field of point clouds, several recent works have achieved promising performance using autoencoders and generative models for unsupervised representation learning of point clouds \cite{method3, ref23,method2}, but they only focus on local structural details and cannot capture higher-level semantic features. Rao et al. \cite{ref26} proposed a global-local bidirectional reasoning approach to consider both local and global information, but it heavily relies on the specific point cloud feature extraction network. Inspired by self-supervised learning in 2D images \cite{ref12,ref13,ref17}, Huang ta al. \cite{method11} utilizes contrastive learning to realize self-supervised point cloud representation learning. However, they do not consider the relationship between local structures and the complete point clouds, leading to limited generalization ability and much difficulty in further performance improvement on downstream tasks with pre-trained models.

This work is inspired by how human beings understand the 3D world. As shown in Fig.\ref{fig1}, humans learn to understand an 3D object by combining two different levels of information. One is the global shape of the object, and the other is its local structure. Inspired by this common sense, we propose a new self-supervised point cloud representation learning framework combining knowledge distillation and contrastive learning. Our method is simple and effective, and does not depend on a specific network architecture. The framework consists of two branches: a teacher network and a student network. First, we use simple random cropping to get global and local point clouds with different sizes. Next, all point clouds are fed into the student network, while only global point clouds are fed into the teacher network. For both the teacher network and the student network, we use existing point-based networks or a ViT \cite{ref37} variant we proposed. The student network directly predicts the output distribution of a teacher network by using a standard cross-entropy loss. Finally, the momentum update in contrastive learning is utilized to update the weight of the teacher network. The updated teacher network can extract better representation and then make the student network improve. Through the training of these steps, the network can learn the way of how human beings understand the 3D world.

\textbf{Our main contributions are as follows:}

1) We integrate knowledge distillation and contrastive learning for the first time to propose a simple and general self-supervised point cloud representation learning framework. In this framework, a point cloud network can learn both the invariance of global shape and the relationship between local structures and the global shape. 

2) We propose a pure transformer point cloud network based on ViT \cite{ref37} (denoted as 3D-ViT), which achieves comparable performance with current state-of-the-art backbones.  Especially, 3D-ViT can help us better analyze what the network has learned. 

3) Our pre-training models achieve the state-of-the-art performance in both linear classifier evaluation and several other downstream tasks.

\section{Related work}
\textbf{Knowledge Distillation} Knowledge distillation is originally introduced to transfer the knowledge of a complex model or multiple ensemble models (teacher network) to another lightweight model, ensuring that the original performance is not lost during the transfer process \cite{ref1, ref2, ref3}. Many researchers further developed knowledge distillation by different optimization objectives, including consistency on feature maps \cite{ref5, ref6}, consistency on probability mass function \cite{ref7}, and maximizing the mutual information \cite{ref4}. Previous work required training a teacher network in a supervised manner. Li et al. \cite{ref9} introduced a self-distillation framework for long-tailed recognition, which trains a self-distillation network under the hybrid supervision of soft labels from previous stages and hard labels from the original. This still relied on partial labeled data. Xu et al. \cite{ref40} extracts structured knowledge for 2D image task from a self-supervised auxiliary task. Mathilde et al. \cite{ref10} further extended the combination of vision transformers and knowledge distillation and achieved better performance. Unlike Mathilde et al. \cite{ref10}, we focus on 3D point clouds, which are extremely different from 2D images. In the point cloud field, it is still an open question whether it is possible to apply distillation to realize knowledge transfer, and whether the combination of knowledge distillation and contrastive learning can work.

\textbf{Contrastive learning} Contrastive learning is a branch of self-supervised learning, which learns knowledge from the data itself without the demand of data annotation. The main idea of contrastive learning is to maximize the consistency between positive sample pairs and the differences between negative sample pairs. Representative methods of contrastive learning include MoCo series \cite{ref11,ref12,ref13} and SimCLR \cite{ref14}. Recently, BYOL \cite{ref17} and Barlow twins \cite{ref18} pointed out that only using positive samples can still obtain powerful features. Our framework is inspired by contrastive learning, but our goal is to continuously improve the performance of the teacher network in knowledge distillation.

\textbf{Point Cloud Self-Supervised Learning} In the past few years, supervised point cloud approaches have made great progress \cite{ref19,ref20,ref21}. However, point cloud annotation is expensive and it is often very costly or even scarce. In practical applications, the performance of supervised approaches is difficult to be further improved. Therefore, self-supervised point cloud representation learning becomes more and more important. Many works have achieved good performance by using autoencoders and generative model to realize self-supervised point cloud representation learning, such as generative adversarial networks (GAN) \cite{ref23}, auto-encoders (AE) \cite{method3, ref25}, and Gaussian mixture models (GMM) \cite{method2}. Most existing methods rely on distribution estimation tasks or reconstruction tasks to provide a supervision, which can only make the network learn better local detail features, but cannot capture higher-level semantic features \cite{ref26}.  To address this issue, Rao et al. \cite{ref26} proposed to learn point cloud representation by bidirectional reasoning between the local structures at different abstraction hierarchies and the global shape without human supervision. Although this method obtains good performance, it relies too much on the specific point cloud feature extraction network. At the same time, it still needs distribution estimation task and reconstruction task to obtain good performance. Inspired by self-supervised learning in 2D images \cite{ref12,ref13,ref17}, Huang et al. \cite{method11} used the idea of contrastive learning to realize self-supervised point cloud representation learning. However, they only focus on the invariant representation between different perspectives, and do not consider the relationship between global shape features and local structure features, leading to limited generalization ability and difficulty in further performance improvement on downstream tasks. Our method also utilizes the idea of contrastive learning. The difference is that our method explicitly learns global shape information and the relationship between the point cloud and its multiple local structures. The extracted features integrating both global and local information can achieve better performance and generalization in different downstream tasks.

\section{Method}
Our framework combines knowledge distillation and contrastive learning, so we denote our framework as \textbf{DCGLR} (combining \textbf{D}istillation and \textbf{C}ontrast to extract \textbf{G}lobal and \textbf{L}ocal \textbf{R}epresentation). For easier understanding of our framework, we will first briefly review knowledge distillation and contrastive learning. Then, we describe how to combine these two techniques for self-supervised point cloud representation learning. Finally, 3D-ViT, a variant of ViT for point cloud feature extraction, is introduced.

\subsection{Preliminaries}
\subsubsection{Knowledge Distillation}
Knowledge distillation is a learning paradigm for knowledge transfer. It consists of two networks: a teacher network $f_{t}$ and a student network $f_{s}$. By restricting the output of the student network as consistent as possible with that of the teacher network, the student network is able to learn the same capabilities as the teacher network. Specifically, given a signal $x$, we can obtain the output of the teacher network $o_{t}=f_{t}(x)$ and the output of the student network $o_{s}=f_{s}(x)$. The teacher network is a fixed pre-trained model, and the student network is updated by minimizing the difference between $o_{t}$ and $o_{s}$, which is defined as,
\begin{equation}
  \setlength{\abovedisplayskip}{2pt}
  \setlength{\belowdisplayskip}{2pt}
  {loss}_{d}=\underset{f_{s}}{min} \langle o_{t}-o_{s} \rangle
  \label{eq1}
\end{equation}
where $\langle \cdot \rangle$ represents any measure of feature difference.

\subsubsection{Contrastive learning}
Contrastive learning is a learning paradigm of unsupervised learning. The key idea is to encourage the representations of positive sample pairs to be close and that of negative sample pairs to be far away. Advanced contrastive learning frameworks usually consist of two networks: an online network $h_{s}$ and a target network $h_{t}$. The parameters of the online network will be updated after the loss is calculated, while there are three ways to update the target network parameters: using the online network’s weight of the last epoch, end-to-end update with online network by back-propagation and momentum-updated encoder. Specifically, given a sample $x$, we first generate a positive sample $x^+$ by data augmentation, and then randomly select another sample from the dataset as negative sample $x^-$. These samples $x^+$ and $x^-$ are processed by the network $h_{t}$, and the sample $x$ is processed by the network $h_{s}$. We update $h_{t}$ and $h_{s}$ by optimizing the following loss function,

\begin{equation}
  \setlength{\abovedisplayskip}{0.5pt}
  \setlength{\belowdisplayskip}{0.5pt}
  \begin{aligned}
  {loss}_{c}=\underset{h_{s}, h_{t}}{min} - \left({score}\left(h_{t}\left(x^{+}\right), h_{s}(x)\right) - \right. \\ \phantom{=\;\;} \left.
  \quad{score}\left(h_{t}\left(x^{-}\right), h_{s}(x)\right)\right)
  \end{aligned}
  \label{eq2}
\end{equation}

where ${score}(\cdot\ ,\ \cdot)$ denotes the similarity of two vectors. We usually choose the inner product of two vectors as the ${score}(\cdot\ ,\ \cdot)$, defined as follows. 

\begin{equation}
  \setlength{\abovedisplayskip}{0.5pt}
  \setlength{\belowdisplayskip}{0.5pt}
  {score}(a, b) = \langle a, b\rangle
  \label{eq3}
\end{equation}

where $a \in R^{1 \times D}$, $b \in R^{1 \times D}$, and $D$ represents the feature dimension.

% figure 2
\begin{figure*}[ht]
  \centering
  \setlength{\belowcaptionskip}{-0.2cm}
  \includegraphics[width=17cm]{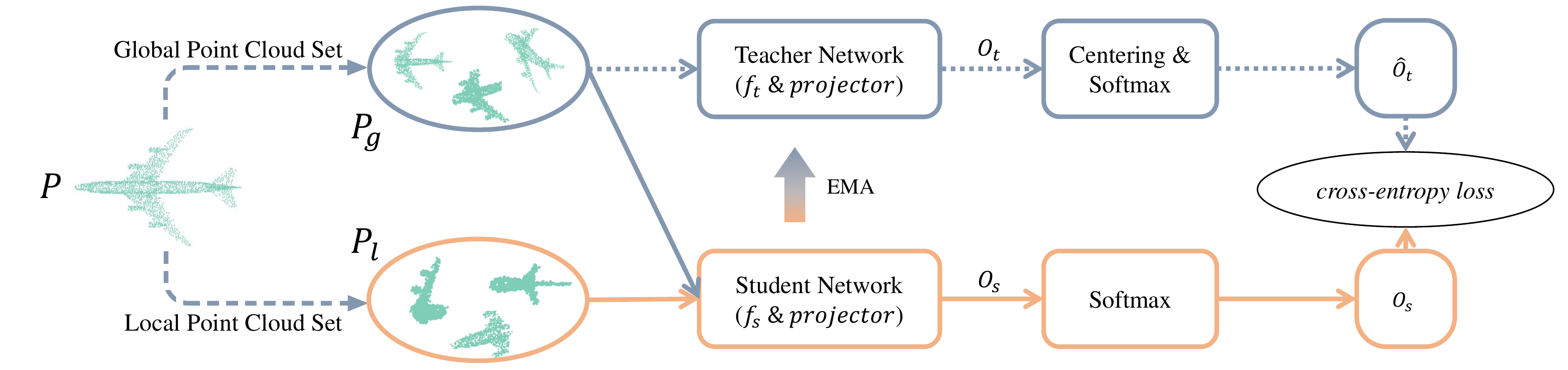} 
  \caption{The overall architecture of DCGLR. Firstly, the global point cloud set and the local point cloud set are obtained by cropping the full point cloud with different cropping ratios. Then, only the global point cloud set is input into the teacher network, but both the global and the local point cloud sets are input into the student network. Finally, the cross-entropy loss between the outputs of the two networks is calculated. EMA represents exponential moving average, solid line represents gradient back-propagation, and dotted line represents stop-gradient operator.}
  \label{fig2}
\end{figure*}

\subsection{DCGLR}
Inspired by the latest self-supervised learning
framework DINO \cite{ref10}, we combine knowledge distillation and contrastive learning for self-supervised point cloud representation learning, so that the backbone network can learn to combine global information and multiple local information to represent point clouds. In Section 3.1, we introduce the basic concepts of knowledge distillation and contrastive learning. In this section, we introduce how to combine them in our framework in detail. As shown in Fig.\ref{fig2}, our framework also consists of two branches, which is similar to the architecture of knowledge distillation and contrastive learning. For convenient description, we first unify names of the two branches. The online network in contrastive learning framework and the student network in knowledge distillation are unified into the same branch and is denoted as student network in our framework. The target network in contrastive learning and the teacher network in knowledge distillation are unified into the same network branch and is denoted as teacher network in our framework. The teacher network and the student network have the same structure but different parameters, and they can by any point-based network, such as PointNet++, DGCNN and etc. 

Different from existing studies, we hope that the network can not only learn the invariance of global shape, but also learn the relationship between local structures and global shape. We utilize knowledge distillation to achieve this goal. Through knowledge distillation, the network can learn the invariance of global shape and the relationship between local structure and global shape at the same time. Specifically, given a input point cloud $P \in R^{N \times 3}$, we generate a global point cloud set $P_{g}$ and a local point cloud set $P_{l}$. The point clouds in $P_{g}$ retain most of the points in the original point cloud $P$, while the point clouds in $P_{l}$ only covers a small area of the original point cloud. The two point cloud sets are generated by cropping the original point cloud as follows.

\begin{equation}
  \setlength{\abovedisplayskip}{0.5pt}
  \setlength{\belowdisplayskip}{0.5pt}
  P_{g}^{i}={crop}\left(P, {rand}\left(r_{g 1}, r_{g 2}\right)\right), \quad i=1 \cdots I
  \label{eq4}
\end{equation}
\begin{equation}
  \setlength{\abovedisplayskip}{0.5pt}
  \setlength{\belowdisplayskip}{0.5pt}
  P_{l}^{j}={crop}\left(P, {rand}\left(r_{l 1}, r_{l 2}\right)\right), \quad j=1 \cdots J
  \label{eq5}
\end{equation}
where ${crop}(\cdot\ ,\ \cdot)$ represents cropping an area at a fixed ratio, represented by the second parameter. ${rand}(\cdot\ ,\ \cdot)$ generates a random value between the maximum and the minimum values. Here, $r_{g1}$ and $r_{g2}$ are the minimum and maximum cropping ratio for generating the global point cloud set, respectively. Similarly, $r_{l1}$ and $r_{l2}$ are the minimum and maximum cropping ratios for generating of the local point cloud set, respectively. $I$ and $J$ are the number of point clouds in $P_{g}$ and $P_{l}$, respectively.

During the training phase, the student network encodes both the global point cloud set and the local point cloud set, while the teacher network only encodes the global point cloud set. There are two reasons for this setting. On one hand, the global point cloud set can get a better representation than the local point cloud set, which can provide more stable optimization objectives for the student network. On the other hand, the student network can be encouraged to learn both the invariance of global shape and local-to-global correspondences. The output of the teacher network and the student network are as follows,

\begin{equation}
  \setlength{\abovedisplayskip}{0.5pt}
  \setlength{\belowdisplayskip}{0.5pt}
  o_{t g}^{i}={ projector }\left(f_{t}\left(\left\{P_{g}^{i}\right\}\right)\right), \quad i=1 \cdots I
  \label{eq6}
\end{equation}

\begin{equation}
  \setlength{\abovedisplayskip}{0.5pt}
  \setlength{\belowdisplayskip}{0.5pt}
  \begin{aligned}
  o_{s g}^{i}, o_{s l}^{j}={ projector }\left(f_{s}\left(\left\{P_{g}^{i}, P_{l}^{j}\right\}\right)\right), \\ i=1 \cdots I, j=1 \cdots J
  \end{aligned}
  \label{eq7}
\end{equation}
where $o_{t g}^{i} \in R^{K}, o_{s g}^{i} \in R^{K}, o_{s l}^{j} \in R^{K}$. $K$ represents the dimension of the output feature. ${projector}(\cdot)$ projects the output of $f_{t}$ or $f_{s}$ into the $K$-dimensional space, which is implemented by MLP. The features used in downstream tasks are the output of the backbone $f_{t}$.

The previous studies of contrastive learning methods rely on large batch size \emph{B} to ensure the stability of training, where B is usually greater than 1000. Inspired by Mathilde et al. \cite{ref10}, we use balanced centering and sharpening on the teacher branch to ensure stable training under a small batch size. The centering dynamically finds a center feature $c$ in the feature space to avoid being dominated by a main direction of the features, encouraging a uniform distribution of features. The features used to calculate the loss is the subtraction between the network output and the center feature $c$, which is defined as follows.

\begin{equation}
  \setlength{\abovedisplayskip}{0.5pt}
  \setlength{\belowdisplayskip}{0.5pt}
  \hat{o}_{t g}^{i}=o_{t g}^{i}-c
  \label{eq8}
\end{equation}

The center feature $c$ is updated by the average of all feature outputs in the batch using the exponential moving average strategy. We calculate the updated center feature $c$ as follows,
\begin{equation}
  \setlength{\abovedisplayskip}{0.5pt}
  \setlength{\belowdisplayskip}{0.5pt}
  c=q * c+(1-q) \frac{1}{B} \sum_{i=1}^{B} o_{tg}^{i}
  \label{eq9}
\end{equation}
where $q>0$ is a rate parameter. The sharpening is implemented through softmax, and temperature $\tau$ controls the sharpness of the output distribution. A lower temperature will result in a sharper output distribution. Our loss includes a global-to-global loss and a local-to-global loss. We minimize the two losses by minimizing the following two cross-entropy,
\begin{equation}
  \setlength{\abovedisplayskip}{3pt}
  \setlength{\belowdisplayskip}{0.5pt}
  { loss }_{{g}}=\min _{f_{s}} \frac{1}{I(I-1)} \sum_{i=1}^{I} \sum_{j=q \& j \neq i}^{I} H\left(\hat{o}_{{tg }}^{i}, o_{s g}^{j}\right).
  \label{eq10}
\end{equation}
\begin{equation}
  \setlength{\abovedisplayskip}{0.5pt}
  \setlength{\belowdisplayskip}{0.5pt}
  {loss}_{{l}}=\min _{f_{s}} \frac{1}{I \cdot J} \sum_{i=1}^{I} \sum_{j=1}^{J} H\left(\hat{o}_{t g}^{i}, o_{s l}^{j}\right)
  \label{eq11}
\end{equation}
where $H$  represents cross-entropy loss, e.g. $H(a, b)=-a \cdot \log b$. The final loss is the weighted average of the two losses.
\begin{equation}
  \setlength{\abovedisplayskip}{1pt}
  \setlength{\belowdisplayskip}{1pt}
  {loss}=\omega_{g} * {loss}_{{g}}+\omega_{l} * {loss}_{{l}}
  \label{eq12}
\end{equation}
where $\omega_{g}$, $\omega_{l}$ are hyper-parameters and equal to 1 in our experiments. 

In knowledge distillation, there is usually a powerful pre-trained model to be used as the teacher network. However, we do not have a pre-trained teacher network in our self-supervised point cloud representation learning. Here, based on the architecture similarity between knowledge distillation and contrastive learning, we utilize the update rules in contrastive learning to dynamically update the teacher network. Grill's research on contrastive learning \cite{ref17} shows that even if the teacher network is randomly initialized, the student network can still learn a better output than a random network. This result provides a strong support for knowledge distillation from global point clouds and local point clouds without a powerful pre-trained teacher network. Therefore, we randomly initialize the teacher network. Although randomly initialized networks can play a role as teachers in the early stages of training, the performance of student networks will stop improving without the guidance of a better teacher network. Accordingly, we need a teacher network that can dynamically update and improve while at the same time the output will not change rapidly before and after each update. Naturally, we adopt momentum updates commonly used in contrastive learning for updating the teacher network, which is defined as follows.

\begin{equation}
  \setlength{\abovedisplayskip}{0.5pt}
  \setlength{\belowdisplayskip}{0.5pt}
  \theta_{t}=\lambda \theta_{t}+(1-\lambda) \theta_{s}
  \label{eq13}
\end{equation}
where $\theta_{t}$ represents the parameters of the teacher network, $\theta_{s}$ represents the parameters of the student network, and $\lambda \in[0,1)$ is a momentum coefficient. 

In short, through knowledge distillation, the student network can learn two kinds of knowledge, the global shape of point clouds and the relationship between local structure and global shape. Through the momentum update rule, we can steadily transfer these two kinds of knowledge that the student network learns to the teacher network, and the updated teacher network provides a stable optimization objective for the student network.

% figure 3
\begin{figure}[t]
  \centering
  \setlength{\belowcaptionskip}{-0.3cm}
  \includegraphics[width=7.5cm]{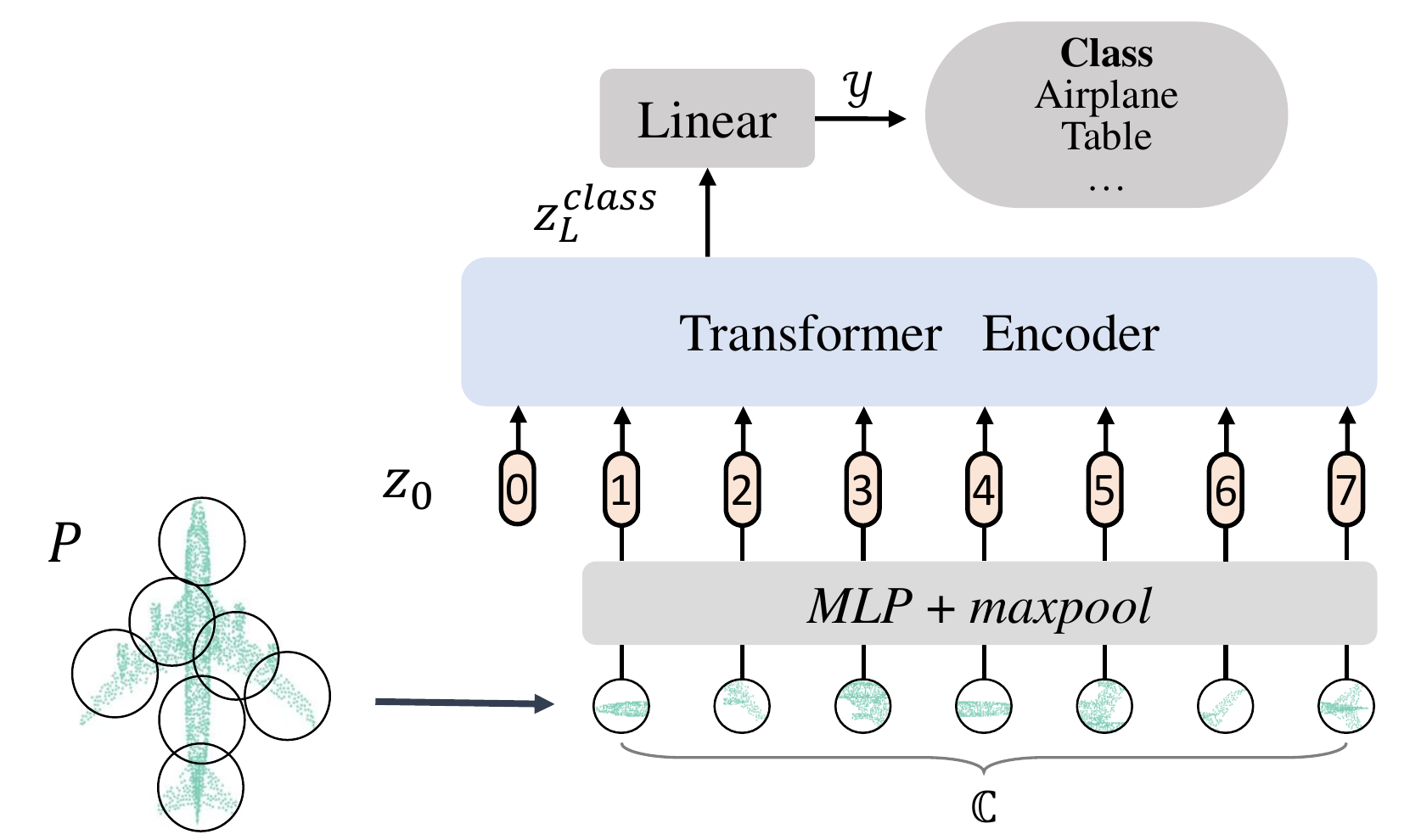} 
  \caption{The framework of 3D-ViT. Firstly, a series of patches $\mathbb{C}$ containing a fixed number of points are extracted from the original point cloud $P$ and the patch embedding token $z$ is obtained by using MLP and maxpooling operation. Then, the sequence $z_{0}$ of class token and patch embedding token is input to the Transformer for encoding. Finally, the class token $z_{0}^{class}$ is used for final classification.}
  \label{fig3}
\end{figure}

\subsection{3D Vision Transformer (3D-ViT)}
The existing point-based networks cannot intuitively analyze what exactly the network learns. Inspired by ViT \cite{ref37} in 2D images, we propose 3D-ViT for point cloud analysis. Due to the big difference in data structure between point clouds and images, 3D-ViT is similar to ViT only in structure but needs very different implementations. 

The overall structure of 3D-ViT is shown in Fig.\ref{fig3}. The input of the standard Transformer \cite{ref38} is a 1D token embedding, so we first extract a series of point cloud patches from the input point cloud $P \in R^{N \times 3}$. The shape of the set of point cloud patches $\mathbb{C}=\left\{P^{i} \mid i=1 \cdots S\right\}$ after grouping is $(S \times k \times 3)$, where $k = 32$ represents the number of points contained in each patch, $S={int}(N / k)$ is the number of patches, and ${int}(\cdot)$ means rounding the scalar value. We adopt the FPS (Farthest Point Sampling) to get the centroids of $S$ patches, and then use knn to get the other $k-1$ points in each patch. The features of each patch are mapped into $D$ dimension by inputting $\mathbb{C}$ into the MLP and performing the maxpooling operation, and these outputs are called patch embedding token $z$. This process is formulated in Eq.\eqref{eq14}. Similar to ViT, we also add a learnable class token $z_{0}^{class}$ and place it before the patch embedding token list. The sequence $z_{0}$ is then input into the standard Transformer, and the corresponding class token value is input into the linear layer for dimension mapping to obtain the final output $y$. 
\begin{equation}
  \setlength{\abovedisplayskip}{0.5pt}
  z_{0}^{i}={maxpool}\left(MLP\left(P^{i}\right)\right), \quad i=1 \cdots S
  \label{eq14}
\end{equation}
\begin{equation}
  z_{0}=\left[z^{class}, z_{0}^{1}, z_{0}^{2}, \cdots, z_{0}^{s}\right]
  \label{eq15}
\end{equation}
\begin{equation}
  \setlength{\belowdisplayskip}{0.5pt}
  y={Linear}\left({Transformer}\left(z_{0}, L\right)[{class token}]\right)
  \label{eq16}
\end{equation}
where $L$ represents the number of layers of multi-head self-attention in the Transformer.

\section{Implementation and Dataset}
\subsection{Implementation}
We use adamw optimizer \cite{ref39} to train the network. The learning rate increases linearly for the first 10 epochs and then decays with a cosine schedule. For the exponential moving average weight $\lambda$ of the teacher network, the starting value is set to 0.996 and then gradually increases to 1. The temperature $\tau$, used to control sharpness of the output distribution, is set to 0.04. The dimension K of the final features used to calculate the loss is set to 512. When cropping the global point cloud, the crop ratios $r_{g1}$, $r_{g2}$ are set to 0.7 and 1.0, respectively, and the number of crops I is 2. When cropping local point clouds, the crop ratios $r_{l1}$, $r_{l2}$ are set to 0.2 and 0.5, respectively, and the number of crops $J$ is 8. Additionally, we use the FPS sample half of the original point cloud as different resolution point clouds and add them to local point cloud set. The number of different resolution point clouds is 2. Our pre-trained model is obtained by training on NVIDIA A100.

\subsection{Dataset}
In the experiments of this paper, three datasets (ShapeNet \cite{ref33}, ModelNet40 \cite{ref31}, and S3DIS \cite{ref34}) are used.
ShapeNet consists of 57,448 synthetic objects from 55 categories. We follow the processing method of Huang et al. \cite{method11}, and each processed point cloud contains 2048 points. Unless otherwise specified, our models are pre-trained on ShapeNet.
ModelNet40 contains 12,331 objects (9,843 for training and 2,468 for testing) from 40 categories. For this dataset, there are two commonly used versions in current researches. The first version follows Qi et al \cite{ref32}, where 10,000 points are sampled directly from the original CAD models, and we denote the first version as ModelNet40\_CAD. In the second version (processed by Stanford University), each point cloud contains 2048 points and is stored in h5py format, and we denote this version as ModelNet40\_H5py. We conducted experiments on ModelNet40 with both existing versions.
S3DIS contains real scan point clouds of 272 rooms scanned in six different regions, and each point is labeled as one of 13 categories as semantic segmentation. We follow Huang et al. \cite{method11} to divide each room into 1m×1m blocks and only use the XYZ coordinates of each point as the model input.

\begin{table}\scriptsize
  \centering
  \setlength{\abovecaptionskip}{-0.1cm} 
  \setlength{\belowcaptionskip}{-0.39cm}
  \caption{Classification results with linear SVM of our method against previous methods on ModelNet40\_CAD}
  \begin{tabular}{lcr}\\
      \hline Method & Input & Accuracy \\
      \hline 3D-GAN \cite{method1} & voxel & $83.3 \%$ \\
      VIP-GAN \cite{method7} & views & $90.2 \%$ \\
      Latent-GAN \cite{method2} & points & $85.7 \%$ \\
      MRTNet \cite{method3} & points & $86.4 \%$ \\
      SO-Net \cite{method4} & points & $87.3 \%$ \\
      FoldingNet \cite{method5} & points & $88.4 \%$ \\
      MAP-VAE \cite{method6} & points & $88.4 \%$ \\
      3D-PointCapsNet \cite{method8} & points & $88.9 \%$ \\
      Sauder et al. + DGCNN \cite{method9} & points & $90.6 \%$ \\
      Poursaeed et al. + DGCNN \cite{method10} & points & $90.7 \%$ \\
      STRL + DGCNN \cite{method11} & points & $90.9 \%$ \\
      STRL + PCT* \cite{method11} & points & $85.3 \%$ \\
      Our + DGCNN & points & $\mathbf{9 1 . 0 \%}$ \\
      \hline
      Our + 3D-ViT & points & $91.3 \%$ \\
      Our + PCT & points & $\mathbf{9 1 . 4 \%}$ \\
      \hline
  \end{tabular}
  \begin{tablenotes}
  \item[1] Note: * indicates that the result is reproduced by us.
  \end{tablenotes}
  \label{tab1}
\end{table}
% figure 4
\begin{figure}[t]
  \centering
  \setlength{\belowcaptionskip}{-0.4cm}
  \includegraphics[width=6cm]{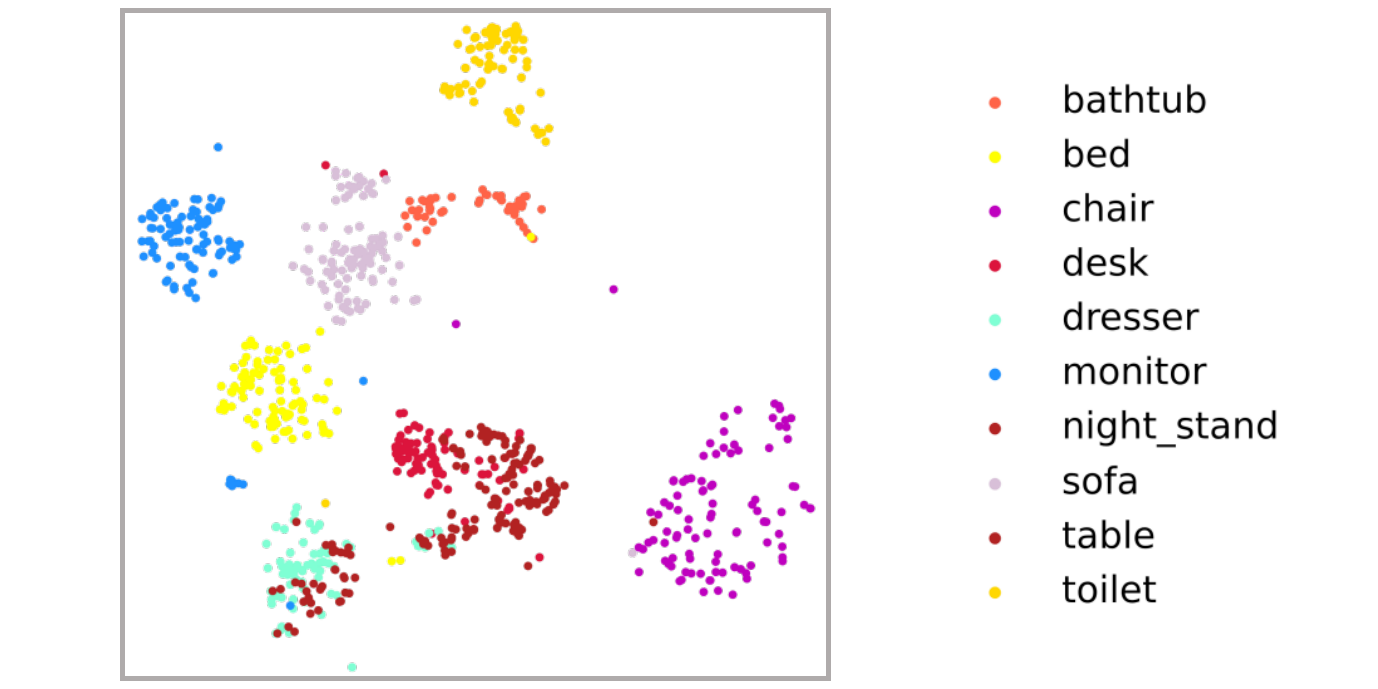} 
  \caption{Visualization of the extracted features in ModelNet10.}
  \label{fig4}
\end{figure}

\section{Experiment}
\subsection{Shape Understanding}
\textbf{Linear SVM}
In this experiment, we compare DCGLR with other methods in extracting features for classifying shapes with linear Support Vector Machine (SVM). For a fair comparison with previous studies, we followed the commonly used setting in \cite{method11} to pre-train the model on ShapeNet and then test it on the ModelNet40\_CAD. The number of point clouds was down-sampled to 2048 points for both training and testing. For DCGLR, we experimented on three backbone point-based networks, DGCNN (a commonly used network in previous studies), a recently proposed transformer-based network PCT \cite{method14}, and our proposed 3D-ViT. In order to classify point cloud shape using the extracted features, we trained a linear SVM on the extracted features of the ModelNet40\_CAD train split and evaluated it on the test split. Table \ref{tab1} shows the classification accuracy of our method and other comparison methods under the SVM classifier. From Table \ref{tab1}, we can see no matter which backbone is used, DCGLR outperforms all existing methods. The best performance is achieved by using DCGLR together with PCT, while the performance of STRL drops sharply when using PCT as the backbone. Meanwhile, a high classification accuracy is also achieved when our 3D-ViT is used. These results demonstrate the high compatibility of DCGLR with a wide range of the backbone networks and they also shows that transformer-based networks are more suitable for large-scale self-supervised point cloud representation learning. The combination result of STRL and PCT is poor, so they will not be combined in subsequent experiments. We visualize the features extracted from the ModelNet10 dataset by DCGLR+DGCNN using T-SNE \cite{ref36} in Fig.\ref{fig4}. It is obvious that the features extracted by our method appear to be clearly separated between different classes, and they are tightly aggregated in the same class, which visually demonstrates the effectiveness of our method in feature extraction.

\textbf{Fine-Tuning} To validate whether our method can help to boost the downstream task performance, we experiment on the point cloud classification task. Table \ref{tab2} shows the results of different classification models under different initialization methods, where, \textbf{from scratch} stands for training the model on ModelNet40\_H5py from randomly initialized network and \textbf{Pretrain} stands for pre-training the model on ShapeNet and then fine-tune the network on ModelNet40\_H5py. The networks initialized by DCGLR pretrain achieve a significant performance improvement over the randomly initialized network. Concretely, our pre-training method improves the classification accuracy 1$\%$ when using DGCNN and 0.5$\%$ when using PCT from training from scratch, helping DGCNN and PCT outperform other supervised methods. With DGCNN as the backbone, our method significantly outperforms the existing state-of-the-art methods \cite{method9,method11}, as shown in lines 7 to 9 of the Table \ref{tab2}.

\begin{table}\scriptsize
  \centering
  \caption{Shape classification results fine-tuned on ModelNet40\_H5py.}
  \setlength{\abovecaptionskip}{-0.25cm} 
  \setlength{\belowcaptionskip}{-0.5cm}
  \begin{tabular}{clc}
    \hline Category & \multicolumn{1}{c}{ Model } & Acc \\
    \hline \multirow{5}{*}{from scratch}  
    & PointNet++ \cite{method12} & $90.7 \%$ \\
    & DGCNN \cite{method13} & $92.2 \%$ \\
    & PCT \cite{method14} & $92.9 \%$ \\
    & PointASNL \cite{method16} & $92.9 \%$ \\
    & Shell \cite{method17} & $93.1 \%$ \\
    & 3D-ViT & $91.6 \%$ \\
    
    \hline \multirow{4}{*}{Pretrain} 
    & Sauder et al. $+$ DGCNN \cite{method9} & $92.4 \%$ \\
    & STRL $+$ DGCNN \cite{method11} & $93.1 \%$ \\
    & Our $+$ DGCNN & $\mathbf{93.2 \%}$ \\
    & Our $+$ 3D-ViT & $92.2 \%$ \\
    & Our $+$ PCT & $\mathbf{93.4 \%}$ \\
    \hline
  \end{tabular}
  \label{tab2}
\end{table}

\subsection{Semantic Segmentation}

This experiment is designed to evaluate whether our framework can effectively improve the semantic segmentation performance of the existing methods when it is used as a pre-training method. Semantic segmentation is done on the Stanford Large-Scale 3D Indoor Spaces (S3DIS) dataset. Following Huang et al \cite{method11}, we used only XYZ coordinates of 4096 points as model input, and we fine-tuned the pre-trained model in one region at a time in Area 1-5, and test the model in Area 6. Obviously, the domain gap between the dataset used for pre-training and the target dataset influences the effectiveness of the pre-training. However, we hope to pre-train a model with good generalizability, which is a hard but more practical task. Therefore, in this experiment, we still used the synthetic dataset ShapeNet to pre-train DGCNN network. STRL pre-trains DGCNN on real dataset ScanNet. We compare the results of STRL \cite{method11} reported in their original paper. The results of this experiment are shown in Table \ref{tab3}. Our method achieves a significant performance improvement compared to the results of training from scratch even though there is a large domain gap between the pre-training dataset and the downstream task dataset. Our approach on the large-gap pre-trained dataset outperforms STRL on the small-gap pre-trained dataset beyond our expectation. This can fully demonstrate that the pre-trained model obtained by our framework has better representation ability and stronger generalization ability.

\begin{table}\scriptsize
  \centering
  \caption{Semantic segmentation results when fine-tuned on S3DIS.}
  \setlength{\belowcaptionskip}{-0.5cm}
  \begin{tabular}{cccc}
    \hline Fine-tuning Area & Method & Acc & mIoU \\
    \hline \multirow{3}{1in}{Area 1 (3687samples)} 
    & from scratch & $84.57 \%$ & $57.85$ \\
    & STRL \cite{method11} & $85.28 \%$ & $59.15$ \\
    & Our & $\mathbf{8 5 . 3 3 \%}$ & $\mathbf{6 0 . 6 1}$ \\
    \hline \multirow{3}{1in}{Area 2 (4440samples)} 
    & from scratch & $70.56 \%$ & $38.86$ \\
    & STRL\cite{method11}  & $72.37 \%$ & $39.21$ \\
    & Our & $\mathbf{7 3 . 1 1 \%}$ & $\mathbf{4 0 . 0 1}$ \\
    \hline \multirow{3}{1in}{Area 3 (1650samples)} 
    & from scratch & $77.68 \%$ & $49.49$ \\
    & STRL \cite{method11} & $\mathbf{7 9 . 1 2 \%}$ & $51.88$ \\
    & Our & $79.06 \%$ & $\mathbf{5 1 . 9 8}$ \\
    \hline \multirow{3}{1in}{Area 4 (3662samples)} 
    & from scratch & $73.55 \%$ & $38.50$ \\
    & STRL \cite{method11} & $73.81 \%$ & $39.28$ \\
    & Our & $\mathbf{7 4 . 0 5 \%}$ & $\mathbf{3 9 . 3 8}$ \\
    \hline  \multirow{3}{1in}{Area 5 (6852samples) }
    & from scratch & $76.85 \%$ & $48.63$ \\
    & STRL \cite{method11} & $77.28 \%$ & $49.53$ \\
    & Our & $\mathbf{7 8 . 2 4 \%}$ & $\mathbf{5 0 . 2 2}$ \\
    \hline
  \end{tabular}
  \label{tab3}
\end{table}

% figure 5
\begin{figure}[t]
  \centering
  \setlength{\belowcaptionskip}{-0.39cm}
  \includegraphics[width=8.3cm]{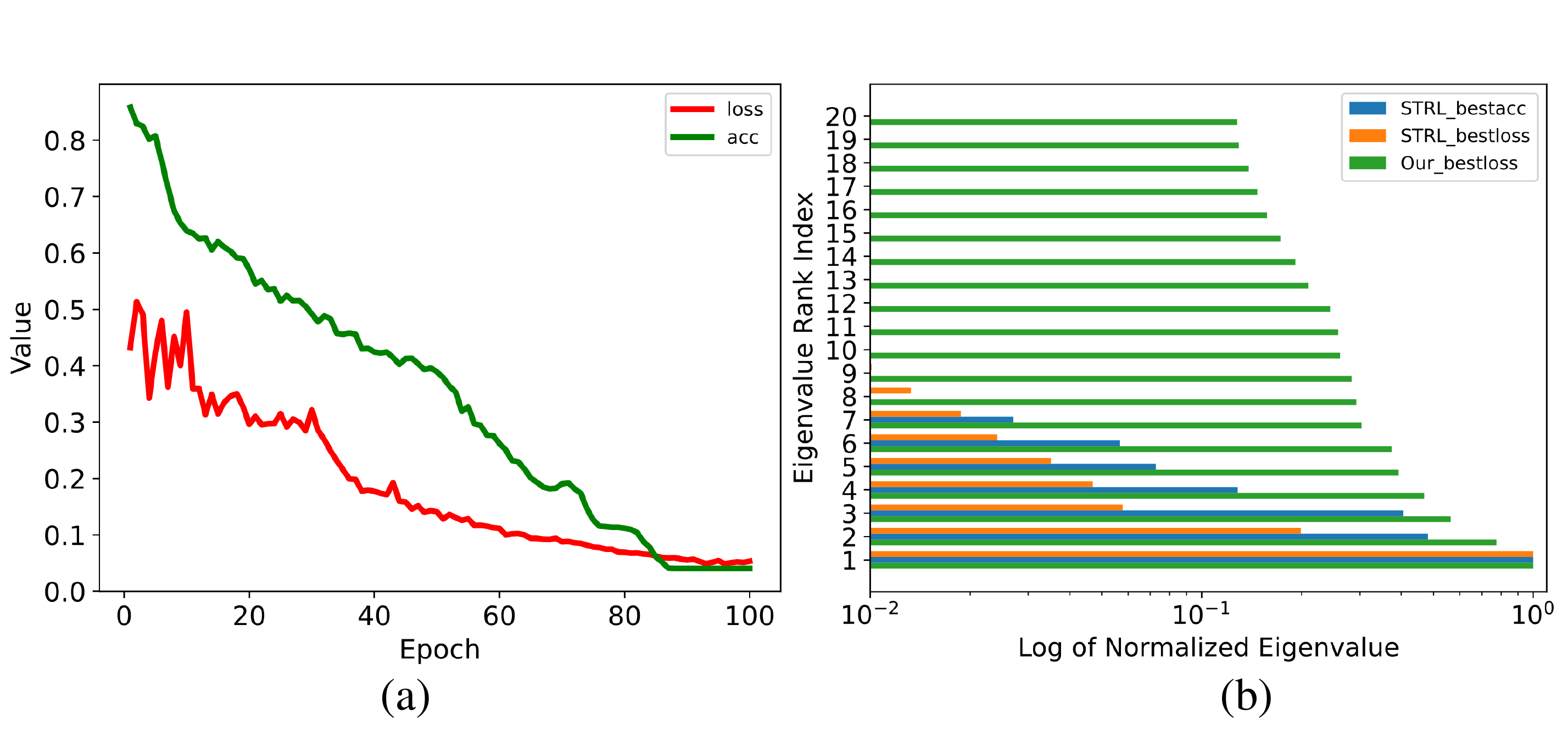} 
  \caption{(a) Loss and acc of the STRL framework in training PCT, (b) Eigenvalues distribution of the output features in PCT trained by each method}
  \label{fig5}
\end{figure}

% figure 6
\begin{figure*}
  \centering
  \setlength{\belowcaptionskip}{-0.39cm}
  \includegraphics[width=17.9cm]{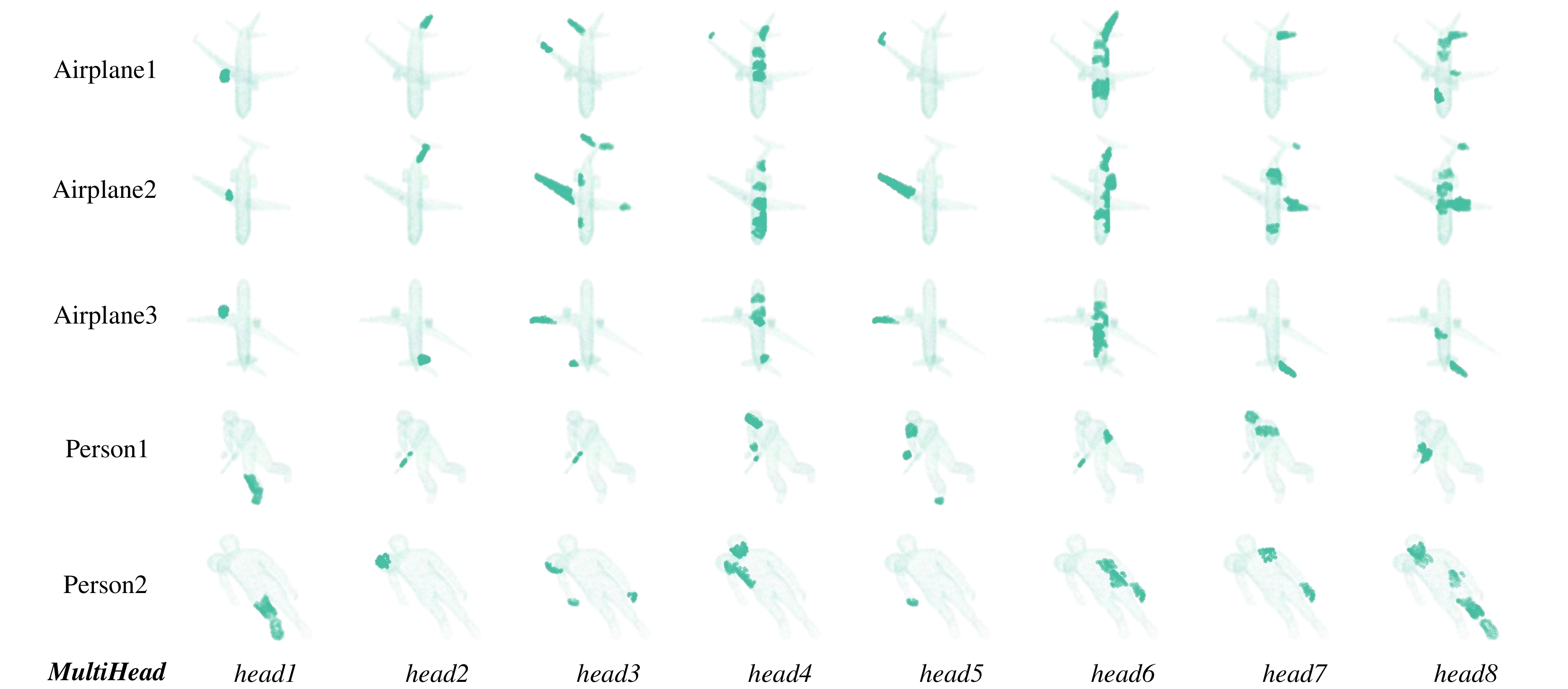} 
  \caption{Attention map visualization}
  \label{fig6}
\end{figure*}

\subsection{Model robustness}
This experiment is designed to further validate the robustness of our method and compare it to the latest point cloud representation learning method STRL \cite{method11}. We used DGCNN as the backbone network in this experiment and pre-trained it using STRL or our DCGLR. Then we investigated the linear classification performance of the pre-trained model using SVM classifier. The model was trained with different numbers of points of the ShapeNet dataset (2048 and 1024), and validated on ModelNet40\_CAD and ModelNet40\_H5py with different point numbers (2048 and 1024). The experimental results are shown in Table \ref{tab4}. The model pre-trained by our method is robust under different settings, consistently outperforming or on par with the previous state-of-the-art method STRL.

\begin{table}\scriptsize
  \centering
  \caption{Shape classification comparison when fine-tuned on ModelNet40\_H5py.}
  \setlength{\abovecaptionskip}{-0.25cm} 
  \setlength{\belowcaptionskip}{-0.5cm}
  \begin{tabular}{lcccc}
    \hline \diagbox{test data}{train data} & \multicolumn{2}{c}{ ShapeNet (2048) } & \multicolumn{2}{c}{ ShapeNet (1024) } \\
    \hline & STRL & Our & STRL & Our \\
    \hline ModelNet40\_CAD(2048) & $90.8 \%$ & $\mathbf{9 1 . 0 \%}$ & $90.2 \%$ & $\mathbf{9 0 . 4 \%}$ \\
    ModelNet40\_CAD(1024) & $\mathbf{8 9 . 9 \%}$ & $\mathbf{8 9 . 9 \%}$ & $\mathbf{9 0 . 0 \%}$ & $\mathbf{9 0 . 0 \%}$ \\
    ModelNet40\_H5py(2048) & $90.2 \%$ & $\mathbf{9 1 . 0 \%}$ & $89.9 \%$ & $\mathbf{9 0 . 9 \%}$ \\
    ModelNet40\_H5py(1024) & $89.7 \%$ & $\mathbf{9 0 . 8 \%}$ & $90.3 \%$ & $\mathbf{9 0 . 7 \%}$ \\
    \hline
  \end{tabular}
  \label{tab4}
\end{table}

\subsection{PCT Feature Analysis}
In the previous experiments, we found that STRL showed poor performance in training PCT. Therefore, we carefully analyzed the reasons in this experiment. Fig.\ref{fig5} (a) shows the curves of loss and accuracy (Acc) of STRL in training PCT. We can find that the network loss decreases during the training process, but the Acc also decreases continuously and finally approaches 0. We calculate the covariance matrix $C$ for the output features and examine its eigenvalues \cite{ref41}. We normalize all the eigenvalues by dividing them with the largest one and scale logarithmically. The histogram of the log of normalized eigenvalue is illustrated in Fig.\ref{fig5} (b). The corresponding histogram of our method is also shown in Fig.\ref{fig5} (b) for comparison. We can see that the model trained with STRL suffers from severe feature collapse, while our model has a smooth distribution in several dimensions, which ensures that the output features of our model have good distinguishability.

\subsection{Attention Map Visualization of 3D Vision Transformer}
The purpose of this experiment is to explore what exactly the self-supervised point cloud representation network is learning and whether it is consistent with our intuitive ideas. We used 3D-ViT as the backbone network and visually analyze attention map of its eight heads. Fig.\ref{fig6} shows the results of five point clouds, in which each row represents a point cloud and each column represents the attention weight of one head of class token. We can easily find that when the network analyzes the same class of objects, a specific head always pays attention to the same kinds of local structure. For example, when analyzing the aircraft, the first head always places its attention on the left engine, which indicates that the network learns that the engine is one of the key local structures to determine the class of the aircraft. Moreover, we can also find that the model trained by our method has two characteristics: first, different attention heads focus on different local structures of the object; second, some attention heads focus on multiple parts of the object as a whole. This shows that the model learns the ability to understand objects by extracting both local structure features and global shape features of the point cloud.

\section{Conclusion}
In this paper, we propose a novel framework for self-supervised representation learning of point clouds. This framework helps a point-based network understand point clouds through global shape and multi-local structures by knowledge distillation and contrastive learning. Our method achieves the best results on both linear classifiers and multiple other downstream tasks, including fine-tuned point cloud classification and point cloud semantic segmentation. The framework is also more robust in working with different point-based networks. Meanwhile, we experimentally demonstrate that our method is indeed capable of learning global shape and multi-local structures. We will further explore the performance of our framework on more downstream tasks.

%%%%%%%%% REFERENCES
{\small
\bibliographystyle{ieee_fullname}
\bibliography{egbib}
}

\end{document}